\documentclass[twoside,11pt]{article}

\usepackage{jmlr2e}
\usepackage{mathtools}
\usepackage{graphicx}
\usepackage{epsfig}
\usepackage{subfigure}

\ShortHeadings{Predictive Maintenance Decision Support System}{Nalbantov, Zografov, Georgiev, Todorov, and Bojilova}
\firstpageno{1}

\begin{document}

\title{Predictive Maintenance Tool for Non-Intrusive Inspection Systems}

\author{\name Georgi Nalbantov \email georgi.nalbantov@danlex.bg \\
	\name Dimitar Todorov \email dimitar.todorov@danlex.bg \\
	\name Nikolay Zografov \email nikolay.zografov@danlex.bg \\
	\name Stefan Georgiev \email stefan.georgiev@danlex.bg \\
	\name Nadia Bojilova \email nadia.bojilova@danlex.bg \\
       \addr Danlex EOOD, Bulgaria}

\editor{}
\today

\maketitle

\begin{abstract} 
Cross-border security is of topmost priority for societies. Economies lose billions each year due to counterfeiters and other threats. Security checkpoints equipped with X-ray Security Systems (NIIS-Non-Intrusive Inspection Systems) like airports, ports, border control and customs authorities tackle the myriad of threats by using NIIS to inspect bags, air, land, sea and rail cargo, and vehicles. The reliance on the X-ray scanning systems necessitates their continuous 24/7 functioning being provided for. Hence the need for their working condition being closely monitored and preemptive actions being taken to reduce the overall X-ray systems downtime. In this paper, we present a predictive maintenance decision support system, abbreviated as PMT4NIIS (Predictive Maintenance Tool for Non-Intrusive Inspection Systems), which is a kind of augmented analytics platforms that provides real-time AI-generated warnings for upcoming risk of system malfunctioning leading to possible downtime. The industrial platform is the basis of a 24/7 Service Desk and Monitoring center for the working condition of various X-ray Security Systems.
\end{abstract}


\section{Introduction}

In the past decades, the world economies have faced unprecedented threats to their security and financial interests. It is estimated that up to USD 870 billion, or 1.5\% of global GDP (\cite{OECD2016}) are lost annually due to counterfeit, illegal trafficking of goods, and other threats. The main means for detecting various threats is invariably X-ray Security Systems, which are used for scanning bags, air, land, sea and rail cargo, and vehicles. Other, mostly concomitant, choices for detection are goods profiling, using meta-data for risk calculation, and others.

The predominant reliance on the X-ray scanning systems necessitates their continuous 24/7 functioning being provided for. Hence the need for their working condition being closely monitored and preemptive actions being taken to reduce the overall X-ray systems downtime. 

In this paper, we present a predictive maintenance decision support system, abbreviated as PMT4NIIS (Predictive Maintenance Tool for Non-Intrusive Inspection Systems), which is a kind of augmented analytics platforms that provides real-time AI-generated warnings for upcoming risk of system malfunctioning leading to possible downtime. The industrial platform is the basis of a 24/7 Service Desk and Monitoring center for the working condition of various X-ray systems. To the best of our knowledge, this is the first predictive maintenance solution for non-intrusive inspection systems for non-medical purposes.

The paper is organized as follows. Section \ref{sec:description} presents a description of the predictive maintenance platform, including IT infrastructure, security considerations, and AI predictive models for systems' health status and risk of forthcoming malfunctions. Section \ref{sec:useCase} gives examples of real-time AI-model predictions, both short-term and long-term, which are brought to the attention of X-ray system service specialists. Section \ref{sec:discussion} discusses implications for ends users such as airports, national customs agencies, and X-ray system maintenance service providers. Section \ref{sec:conclusion} concludes.

\section{Description of the predictive maintenance platform for X-ray systems}
\label{sec:description}

 \begin{figure*}[h]
 	\center
 	\hspace*{-0cm}
		\hbox{\hspace{-1.3cm} \includegraphics[height=13.5cm]{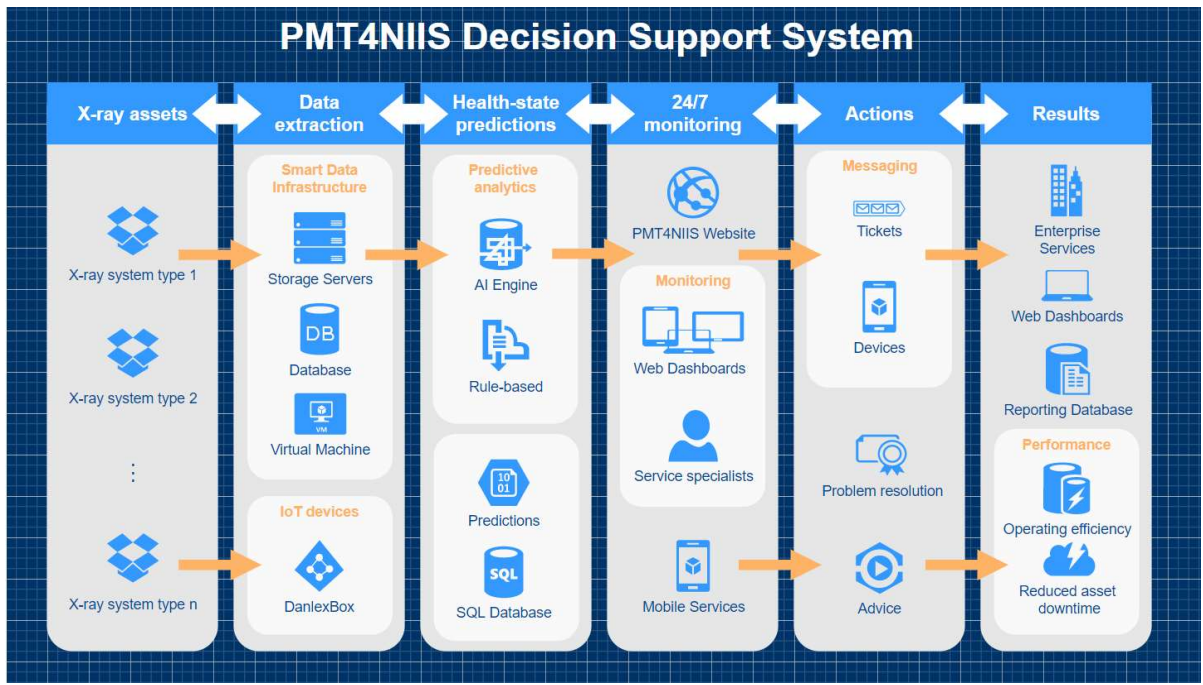}}
 	\vspace*{-4cm}
	\caption{PMT4NIIS Decision Support System design.}
	\label{fig:PMT4NIIS}
 \end{figure*}

The structure of the predictive maintenance tool for non-intrusive (X-ray based) inspection systems (PMT4NIIS) consists of the following elements: an IoT device for extracting relevant technical data from X-ray systems, a Smart Data Infrastructure for secure data transfer and data collection, an AI Engine for generating alerts in case of short-term and/or long-term predictions for high risk of X-ray systems malfunction, and a front-end BI dashboard for visualization of current system parameters as well as AI-generated alerts and warnings. The users of the BI dashboard are mainly service specialists, who receive the output of the decision support system and take an informed decision on whether to intervene and perform system maintenance or do not take any action. Other users of the BI dashboard are end users (airports, customs agencies, port administrations, etc.), who are interested in monitoring the health state and availability of their (X-ray) assets. The elements of PMT4NIIS who they are organized in one overall system are depicted in Figure \ref{fig:PMT4NIIS}.

The assets, which state is being monitored, are various types of industrial non-medical X-ray systems. For the airports, these are predominantly low-energy X-ray imaging systems for carry-on luggage, hand-baggage, as well as air-cargo goods. For border customs, these are high-energy systems that generate X-ray images for trucks and trains content check-up. And for ports, these are high-energy imaging systems for containers. All X-ray systems are subject to wear-and-tear, and therefore require parts to be replaced, as they break and/or malfunction. The traditional maintenance service models are based on preventive (performed at predetermined intervals) and corrective (run-until-it-breaks) maintenance approaches. Both of these services are not efficient (\cite{CXPgroupSummary2018}) and in many industries they are currently being replaced by Condition Based Maintenance (CBM) and Predictive Maintenance (PdM), which ensure the reduction in breakdowns, reduction in maintenance cost and increase of the production (\cite{RolandBerger2014}). What we propose is a shift from the traditional models into Service 4.0 models, which involve real-time asset monitoring and AI-based asset-health alert system, which help anticipate upcoming malfunction problems and consequently prepare for their repair in advance. These actions both save operating costs and increase X-ray systems availability, which are two crucial KPIs.

The data-extraction element of PMT4NIIS involves extraction of machine log files generated by the X-ray systems, which provide relevant data about current technical state. In case additional technical data is needed for better monitoring, which is not readily available in the log files, we have introduced a multi-purpose IoT device, called DanlexBox, for automatic collection and transfer of additional technical and environmental data. The DanlexBox is interoperable across interfaces and system boundaries with autonomous AI-driven functions for processing and cleaning data. All technical and environmental data is collected and transfered in (near) real time using a private APN channel via the DanlexBox IoT device. An alternative approach is to store the data on the IT infrastructure of the owner of X-ray scanners, and then transfer the data from there. All technical data is transferred to a centralized location (outside the IT infrastructure of the owner of X-ray scanners) with a Smart Data Infrastructure, where it is stored and delivered to a predictive analytics environment for post-processing.  DanlexBox is a cyber resilient device supporting the Open Architecture for Airport Security Systems initiative (\cite{Kramer2020}). 

The  predictive analytics environment takes as input raw technical data from X-ray systems. Based on them, it generates and provides predictions for the current health state of the systems. In essence, an AI engine, equipped with AI prediction models, provides short-term (to up a month) and long-term (survival) estimates for the risk of X-ray system stoppage/malfunction. In such a way, both short-term reparation actions as well as long-term inventory management activities can be planned in advance.
The AI models are augmented with rule-based physical models, which generate alarms based on predefined thresholds of certain censor and other technical readings (including error messages, electrical parameters, etc.). All predictions are stored in an SQL database.

The relevant technical X-ray-system data, available either via the machine log or collected via the DanlexBox IoT device, together with the asset health-state predictions, are brought to the attention to service specialists in a 24/7 monitoring center. The service specialists monitoring center are responsible for acting upon generated alarms, where the possible actions are: refute the alarm, accept the alarm, take relevant action based on accepted alarm, and escalating alarm to a higher level. The output of the monitoring center is: 24/7 surveillance of the health-state of X-ray systems, report-generation in case of maintenance action is recommended/advised by the service specialists, and generation of knowledge-base for solved cases. The service specialists can be alerted not only via the available monitoring web dashboards of the PMT4NIIS web site but also via email and mobile phone notification.

 \begin{figure*}[h]
 	\center
 	\hspace*{-0cm}
		\hbox{\hspace{-1.3cm} \includegraphics[height=12cm]{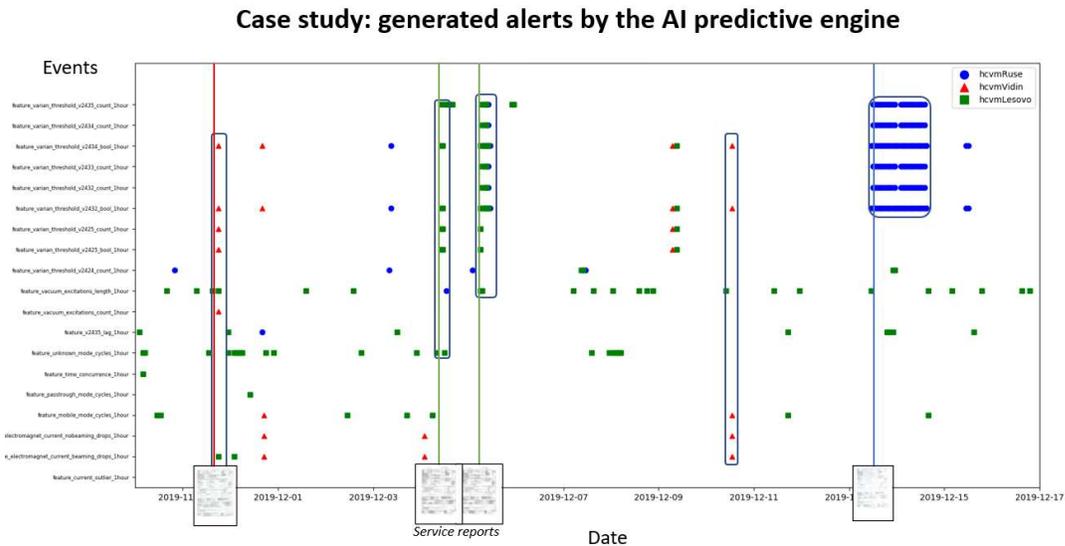}}
 	\vspace*{-4cm}
	\caption{An example use case of the PMT4NIIS decision support system. AI-based generated alerts (colored) as well as associated system-data events are being displayed in a web front end for a chosen period of time (1 dec 2019 - 17 dec 2019). These alerts are being reviewed by maintenance service specialists and physicists and further action is taken where necessary.}
	\label{fig:useCase}
 \end{figure*}

The overall output of the PMT4NIIS decision support system to the X-ray owners and X-ray systems maintenance service providers are (i) web dashboards to check overall system availability and health state, (ii) recommended actions in case of high risk of system malfunction, and (iii) access to ever-increasing knowledge base for solving malfunction problems. The performance of the decision support system is measured in terms of improved operating-efficiency and reduced X-ray systems downtime, resulting in improved financial/economic KPIs.

\section{Use case}
\label{sec:useCase}

In this section we present an example use case of the PMT4NIIS decision support system (see Figure \ref{fig:useCase}). The period in question is 1 dec 2019 - 17 dec 2019. The X-ray systems being monitored are high-energy systems located at cross-border checkpoints \footnote{We do not disclose the location of these systems for security reasons.}. In the case at hand, four technical service reports, which are based on on-site visits, are being provided and used by service specialists to augment the knowledge base for problem resolution management. Various sorts of alerts are being generated through time and brought to the attention of the service specialists at the 24/7 monitoring center. The associated (log) events on which the alerts are being based are displayed as well. In many cases the AI predictive engine generates alerts based on a collection of events, each one participating with its relative weight computed by an AI model. At other times, one event is sufficient to generate an alert, for example when the alert is rule-based (referred to as condition-based alert).

For each alert, the service specialists at the monitoring center decide whether to accept the alert or label it as a false alarm. In this case the alert is labeled as a false alarm, it is provided as feedback to the AI engine, which takes it into account in a re-training process. In such a way, the predictive models increase their performance with time. If an alert is accepted, it is either tackled by the service specialists at hand or escalated to a higher level. A technical report is being generated for each solved case. The alerts generated in this case study have helped the service specialists diagnose an X-ray system \emph{remotely}, and in a timely fashion, without losing any travel time to the location of the X-ray system.

\section{Discussion}
\label{sec:discussion}

We have presented the first, to our best knowledge, an online decision support system for remote monitoring of X-ray non-intrusive inspection systems. It's main features include a secure transfer of machine-log data from X-ray systems to a centralized location with Smart Data Infrastructure. The data is being provided to an AI engine, which applies continuously short-term and long-term models for X-ray systems health state. In case of alerts being generated, they are displayed on a dedicated web-based BI dashboard. The BI dashboard is a web site, where all relevant system data is displayed in near-real time, together will all AI-generated alerts. A monitoring center, which consists of service specialists, takes care of handling the alerts. In case of detected or predicted problems, preemptive action is being taken.

Similar kinds of online decision support systems are already successfully applied in a large variety of other industries. \cite{DBLP:conf/icphm/CanizoOCCT17} present a predictive maintenance AI-based solution for real-time prediction of wind turbines failures. \cite{Olesen2020} describe a data-driven predictive maintenance approach for pump systems and thermal power plants based on a variety of technical inputs. \cite{Massaro2018ESBPI} suggest a predictive maintenance approach for milk production lines, based on inputs such a temperature. By collecting valuable data generated by sensors, using IoT devices for data collection and cloud computing for analyzing data \cite{coban2018} have proposed a predictive maintenance architecture for  medical equipment for improved health-care services.

\section{Conclusion}
\label{sec:conclusion}

It is important to societies to have the technology to inspect, in a non-intrusive way, goods at locations such as airports, cross-border checkpoints, and ports. Such technology exists in the form of X-ray non-intrusive inspection systems. These systems need to be operational at all times, which is currently a bottleneck, which needs to be addressed. To help ensure the smooth functioning of these systems at all times, we have developed a comprehensive decision support system, which reduces the reaction time in case a problem is detected, and also anticipates short-term and long-term problems by employing advanced AI models. A 24/7 monitoring center is indispensable for quicker diagnostic and reaction times, which ultimately results in increased X-ray system availability and improved operational efficiency. The long term effectiveness of this approach vis-a-vis the predominant off-line, run-to-failure and regular-time maintenance approaches should be further researched to provide insights into way to improve industrial and financial KPIs.

\vskip 0.2in
\bibliography{ref}

\end{document}